%% file: main.tex
\documentclass[letterpaper, 10 pt, conference]{ieeeconf}
\IEEEoverridecommandlockouts

\input{usepackages.tex}
\input{macros.tex}

\newcommand{\ctrlseq}{\mathbf{u}}
\newcommand{\vfunc}{V}
\newcommand{\auxvfunc}{\hat{\vfunc}}
\newcommand{\param}{\theta}
\newcommand{\learnedauxvfunc}{\auxvfunc_{\param}}

\newcommand{\learnedpolicy}{\pi_{\param}}

\def\BibTeX{{\rm B\kern-.05em{\sc i\kern-.025em b}\kern-.08em
    T\kern-.1667em\lower.7ex\hbox{E}\kern-.125emX}}

\title{MAD-PINN: A Decentralized Physics-Informed Machine Learning Framework for Safe and Optimal Multi-Agent Control}
\author{ 
Manan Tayal*$^{1}$, Aditya Singh*$^{2}$, Shishir Kolathaya$^{1}$, Somil Bansal$^{3}$
\thanks{$^*$ Denotes equal contribution}
\thanks{$^{1}$Authors are with the Center for Cyber-Physical Systems, Indian Institute of Science, Bangalore, India
(email:\{manantayal, shishirk\}@iisc.ac.in)}
\thanks{$^{2}$Author is with the Department of Electrical and Systems Engineering, University of Pennsylvania, Philadelphia, USA
(email:\{singhadi@engineering.upenn.edu\})}
\thanks{$^{3}$Author is with the Department of Aeronautics and Astronautics, Stanford University, USA (email:\{somil@stanford.edu\})}
}

\begin{document}

\maketitle

\begin{abstract}
Co-optimizing safety and performance in large-scale multi-agent systems remains a fundamental challenge. Existing approaches based on multi-agent reinforcement learning (MARL), safety filtering, or Model Predictive Control (MPC) either incorporate safety via soft constraints that may not reliably prevent unsafe interactions, exhibit conservative performance, or fail to scale effectively. We propose \textit{MAD-PINN}, a decentralized physics-informed machine learning framework for solving the multi-agent state-constrained optimal control problem (MASC-OCP). Our method leverages an epigraph-based reformulation of SC-OCP to simultaneously capture performance and safety, and approximates its solution via a physics-informed neural network. Scalability is achieved by training the SC-OCP value function on reduced-agent systems and deploying it in a decentralized fashion, where each agent relies only on local observations of its neighbors for decision-making. To further enhance safety and efficiency, we introduce a Hamilton-Jacobi (HJ) reachability-based neighbor selection strategy to prioritize safety-critical interactions, and a receding-horizon policy execution scheme that adapts to dynamic interactions while reducing computational burden. Experiments on multi-agent navigation tasks demonstrate that MAD-PINN achieves superior safety–performance trade-offs, maintains scalability as the number of agents grows, and consistently outperforms state-of-the-art baselines. Videos demonstrating the experimental results are provided in \href{https://drive.google.com/drive/folders/1YR76AyYc-YuBClnH7kDQlXLHk13y2VBS?usp=sharing}{the supplementary material}.

\end{abstract}

  

\section{Introduction}
\label{section: introduction}
\input{1_intro_}

\section{Problem Setup} 
\label{section: problem-setup}

\input{2_prob_setup}


\section{Methodology}
\label{section: method}
\input{3_method}

\section{Experiments}
\label{section: experiments}
\input{4_experiments}

\section{Conclusion and Future Work}
\label{section: conclusions}
\input{5_conclusion}

\bibliographystyle{IEEEtran}
\bibliography{ref.bib}

\end{document}

%% file: usepackages.tex
\usepackage[colorlinks]{hyperref}
\hypersetup{
    colorlinks,
    linkcolor=black,
    citecolor=black,
    filecolor=magenta,
    urlcolor=cyan,
}

\usepackage{resizegather}

\usepackage{cite}
\usepackage{amsmath,amssymb,amsfonts,mathrsfs}
\usepackage{mdframed}
\usepackage{graphicx}
\usepackage{textcomp}
\usepackage{xcolor}
\usepackage{tcolorbox}
\usepackage{textcomp}
\usepackage{tablefootnote}
\usepackage{threeparttable}
\usepackage{caption, subcaption}
\usepackage{bbm}
\usepackage{dsfont}
\usepackage{tikz}
\usepackage{graphicx}
\usepackage{tkz-euclide}
\usepackage{algorithm}
\usepackage{algpseudocode}
\usepackage{enumerate}
\usepackage{mathtools}
\usepackage{balance}
\usepackage{booktabs}
\usepackage{titlesec}
\usepackage{flushend}

\usetikzlibrary{positioning}
\usetikzlibrary{shapes}
\usetikzlibrary{shapes.misc}
\usetikzlibrary{shapes.geometric}
\usetikzlibrary{plotmarks}
\usetikzlibrary{intersections}
\usetikzlibrary{calc}
\usetikzlibrary{fit}
\usetikzlibrary{patterns,tikzmark}
\usetikzlibrary{matrix,decorations.pathreplacing,calc}

\tikzset{cross/.style={cross out, draw, 
         minimum size=2*(#1-\pgflinewidth), 
         inner sep=0pt, outer sep=0pt}}


%% file: macros.tex
\newcommand{\vertiii}[1]{{\left\vert\kern-0.25ex\left\vert\kern-0.25ex\left\vert #1 
    \right\vert\kern-0.25ex\right\vert\kern-0.25ex\right\vert}}

\newmdtheoremenv[linecolor=white, backgroundcolor=lightgray!15, innertopmargin=5pt, innerbottommargin=5pt, skipabove=10pt, skipbelow=10pt]{theorem}{\textbf{Theorem}}
\newmdtheoremenv[linecolor=white, backgroundcolor=lightgray!15, innertopmargin=5pt, innerbottommargin=5pt, skipabove=10pt, skipbelow=10pt]{corollary}{\textbf{Corollary}}[theorem]
\newmdtheoremenv[linecolor=white, backgroundcolor=lightgray!15, innertopmargin=5pt, innerbottommargin=5pt, skipabove=10pt, skipbelow=10pt]{lemma}{\textbf{Lemma}}

\newmdtheoremenv[linecolor=white, backgroundcolor=lightgray!15, innertopmargin=5pt, innerbottommargin=5pt, skipabove=10pt, skipbelow=10pt]{problem}{\textbf{Problem}}

\newmdtheoremenv[linecolor=white, backgroundcolor=lightgray!15, innertopmargin=5pt, innerbottommargin=5pt, skipabove=10pt, skipbelow=10pt]{definition}{\textbf{Definition}}

\newmdtheoremenv[linecolor=white, backgroundcolor=lightgray!15, innertopmargin=5pt, innerbottommargin=5pt, skipabove=10pt, skipbelow=10pt]{objective}{\textbf{Objective}}

\newmdenv[
    linecolor=white, backgroundcolor=lightgray!15, innertopmargin=5pt, innerbottommargin=5pt, skipabove=10pt, skipbelow=10pt
]{graybox}

\makeatletter
\renewcommand{\fps@figure}{htp}
\renewcommand{\fps@table}{htp}
\makeatother

%% file: 1_intro_.tex
The deployment of autonomous systems in safety-critical domains such as aerial swarms \cite{soria2021predictive}, intelligent transportation~\cite{donatus2025multiagent} networks, and automated warehouses~\cite{kattepur2018distributed} has made multi-agent coordination a central problem in robotics and control. In these environments, multiple agents must operate in shared spaces and achieve collective objectives, such as routing, formation control, or exploration, while adhering to strict safety constraints. The joint requirement of balancing task performance with safety makes the synthesis of control policies in multi-agent systems a challenging problem.

Multi-agent reinforcement learning (MARL)~\cite{lowe2017multi,yu2022surprising, nayak2023scalable} has emerged as a popular paradigm for policy learning in such settings. While MARL demonstrates strong performance in complex tasks, its safety treatment is insufficient. 
Safety is typically introduced via reward shaping, treating safety constraints as soft penalties.
Constrained Markov Decision Process (CMDP) formulations \cite{altman1999constrained, gu2023safe} provide a more principled approach, but they only ensure that constraint violations remain bounded on average, which is insufficient for safety-critical robotic applications where violations must be avoided at all times. 

Control-theoretic methods such as Control Barrier Function (CBF) \cite{Ames_2017} and Hamilton-Jacobi (HJ) Reachability \cite{10665911, 10266799} provide formal safety guarantees. These methods can act as safety filters \cite{annurev:/content/journals/10.1146/annurev-control-071723-102940} on top of existing nominal controllers, to minimally modify them to enforce safety.
However, they suffer from scalability and conservatism in multi-agent settings.
A popular approach is to compute pairwise safety filters and extend them to many agents; however, this approach remains ineffective, as the intersection of individual safe sets does not necessarily represent the true joint safe set~\cite{choi_aloor_li_2025safemarl}.
\nocite{tayal2024learning}
Moreover, the myopic nature of safety filters often degrades task performance.
Optimal control methods such as Model Predictive Control (MPC)\cite{GARCIA1989335, grune2017nonlinear} and Model Predictive Path Integral control (MPPI)~\cite{8558663, 10161511} provide another line of solutions. 
These methods incorporate predictive look-ahead with explicit handling of hard constraints, and can flexibly handle nonlinear dynamics and diverse cost functions. These approaches have also been extended to multi-agent collision avoidance through the inclusion of modified cost terms or additional constraints \cite{10161511}. Yet, despite their flexibility, MPC and MPPI require solving an optimization problem online at every control step. As the number of agents increases, the resulting optimization becomes increasingly challenging due to the growth in state dimension and inter-agent constraints, limiting real-time applicability in large-scale multi-agent systems.

\begin{figure*}[t]
    \centering
    \includegraphics[width=1.0\textwidth]{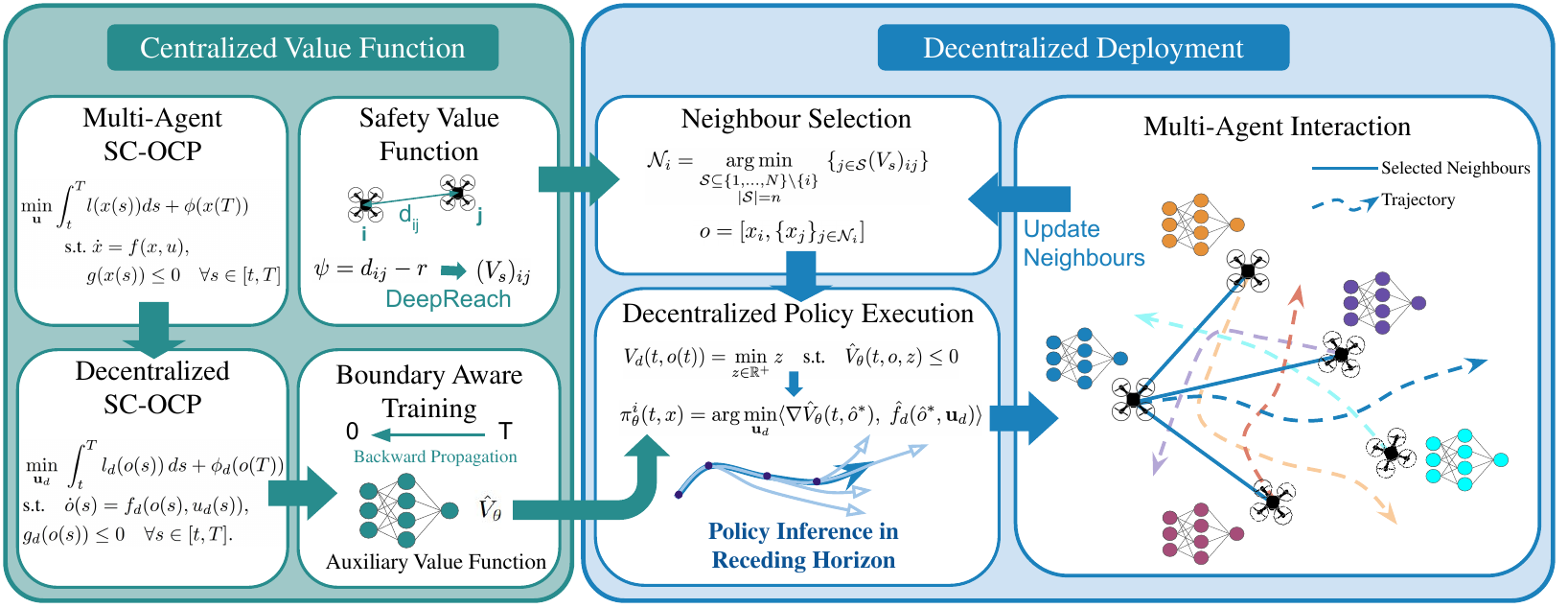}
    \caption{We propose \textbf{MAD-PINN} -- a framework for safe and optimal multi-agent control. 
    MAD-PINN
    is divided into two phases: 
(1) \textbf{Centralized training}, where we learn the auxiliary epigraph-based value function $\learnedauxvfunc$ using a boundary-aware PINN, and the pairwise safety value function $V_s$ via DeepReach; 
(2) \textbf{Decentralized deployment}, where each agent selects its safety-critical neighbours using $V_s$ and executes the policy in a receding-horizon manner using $\learnedauxvfunc$. 
This design enables tractable training, adaptive neighbour selection, and scalable execution in large multi-agent systems.}
    \label{fig:algorithm}
    \vspace{-1.3em}
\end{figure*}

A principled framework for unifying performance and safety is the state-constrained optimal control problem (SC-OCP), which formulates performance as cost minimization and safety as a state constraint. In the multi-agent setting, SC-OCP is particularly appealing because it directly encodes the dual objectives of collision-free coordination and task performance. However, solving SC-OCPs at scale is computationally formidable \cite{doi:10.1137/0324032}. Epigraph-based reformulations recast the problem as a Hamilton-Jacobi-Bellman partial differential equation (HJB-PDE) \cite{altarovici2013general}, but the added dimensionality exacerbates computational complexity. More critically, in centralized multi-agent scenarios, the joint state–action space grows linearly with the number of agents, rendering classical, grid-based numerical PDE solvers impractical even for modest system sizes \cite{8263977}. Thus, while SC-OCP provides a theoretically sound foundation for safe multi-agent control, its direct application to large-scale systems remains infeasible without new strategies for decentralization and scalability.

To address these challenges, we propose \textit{MAD-PINN}, a decentralized physics-informed machine learning framework for solving the multi-agent SC-OCP.  At its core, MAD-PINN combines control-theoretic structure with neural approximation to jointly account for safety and performance while maintaining scalability. Specifically, MAD-PINN uses a physics-informed neural network (PINN) to approximate the epigraph-based value function of the SC-OCP, with boundary conditions encoded to ensure strict satisfaction of terminal safety constraints. This allows MAD-PINN to bypass the computational challenges associated with traditional PDE solvers. Scalability is achieved by training these value functions on reduced-agent systems and deploying them in a decentralized fashion, where each agent makes decisions using only local observations of its most safety-critical neighbors. To support reliable real-world execution, MAD-PINN integrates two additional components: an HJ reachability-based neighbor selection strategy to identify critical interactions, and a receding-horizon policy execution scheme to adapt online to dynamic agent interactions.
Our core contributions are:
\begin{itemize}  
    \item \textbf{MAD-PINN framework:} We propose a \textit{decentralized, boundary-aware} physics-informed learning framework for multi-agent state-constrained optimal control (SC-OCP), which enables the co-optimization of safety and performance in large-scale multi-agent systems.
    \item \textbf{Safety-aware neighbour selection:} We introduce a reachability-based neighbour selection strategy that prioritizes the most safety-critical interactions, thereby preserving safety while avoiding unnecessary conservatism in decentralized execution.  
    \item \textbf{Empirical validation:} We evaluate MAD-PINN on multi-agent navigation tasks at varying scales, showing that it achieves superior safety--performance trade-offs, maintains high scalability as the number of agents grows, and outperforms state-of-the-art baselines.  
\end{itemize} 
%

%% file: 2_prob_setup.tex
Consider a homogeneous multi-agent system comprising $N$ agents. The state and control input of agent $i$ are denoted by $x_i \in \mathcal{X}_i \subseteq \mathbb{R}^{D}$ and $u_i \in \mathcal{U}_i \subseteq \mathbb{R}^{M}$, respectively, with dynamics
$\dot{x}_i(t)=f_i(x_i(t),u_i(t)),$ where $f_i:\mathbb{R}^{D}\times\mathbb{R}^{M}\rightarrow\mathbb{R}^{D}$ is locally Lipschitz continuous. Let $
x=[x_1,\ldots,x_N]\in\mathcal X,~
u=[u_1,\ldots,u_N]\in\mathcal U,$ denote the joint state and control vectors, yielding the global dynamics
$\dot{x}(t)=f(x(t),u(t)).$

The primary safety objective is collision avoidance. We formalize this by defining a failure set $\mathcal{F} \subseteq \mathcal{X}$ of unsafe configurations: $\mathcal{F} = \{ x \in \mathcal{X} : \min_{i \neq j} \, d_{ij}(x) \leq r \}$, where $d_{ij}(x)$ denotes the Euclidean distance between agents $i$ and $j$, and $r$ is a prescribed minimum safe separation distance. Let $g(x)$ be a constraint function such that $ \mathcal F=\{x\in\mathcal X:g(x)>0\}.$

The system's performance is quantified by a cost functional $C(t, x(t), u(\cdot))$ that accumulates a running cost and a terminal cost over a time horizon:
\begin{equation}
    C(t, x(t), u(\cdot)) = \int_{s=t}^{T} l(x(s))  ds + \phi(x(T)),
\end{equation}
where $l: \mathcal{X} \to \mathbb{R}_{\geq 0}$ and $\phi: \mathcal{X} \to \mathbb{R}_{\geq 0}$ are non-negative, Lipschitz continuous functions. The control input is $u(\cdot):[t, T)\rightarrow \mathcal{U}$.
$C$ could represent the time taken to reach the goal or fuel consumption, for instance. Furthermore, we assume the cost functions are separable and identical across the homogeneous agents.  Our goal is to synthesize an optimal control policy $\pi^*: [t, T) \times \mathcal{X} \to \mathcal{U}$ that minimizes this cost while guaranteeing that the state trajectory never enters the failure set $\mathcal{F}$. This leads to the following \textbf{State-Constrained Optimal Control Problem (SC-OCP)} with the value function $V(t, x)$:
\begin{equation}\label{eq: SC-OCP}
    \begin{aligned}
    &V(t, x(t)) = \min_{u(\cdot)}\; \int_t^{T} l(x(s)) ds + \phi(x(T)) \\
    \text{s.t.}~~~ & \dot{x}(s) = f(x(s), u(s)), \quad 
    g(x(s)) \leq 0 ~~ \forall s \in [t, T]
    \end{aligned}
\end{equation}
This SC-OCP enhances the system's performance by minimizing the cost, while maintaining system safety through the state constraint, $g(x) \leq 0$, ensuring that the system avoids the failure set, $\mathcal{F}$. 

The solution of this SC-OCP yields a policy that jointly optimizes safety and performance. As a direct consequence of its formulation,  the centralized SC-OCP in~\eqref{eq: SC-OCP} suffers from the curse of dimensionality; its computational complexity grows linearly with the number of agents $N$ \cite{tayal2025physics}. To address this challenge, each agent reasons over a fixed-size local neighborhood. Specifically, agent $i$ constructs an observation $o_i=\mathcal O_i(x) =
[x_i,\{x_j\}_{j\in\mathcal N_i}],$ where $\mathcal N_i$ contains at most $n$ neighboring agents.
Since $n$ is fixed and independent of the total number of agents, the observation dimension remains constant regardless of system size.

Owing to the homogeneity of the agent dynamics and objectives, all agents share an identical decentralized value function. Dropping the agent index, the resulting \textbf{Decentralized SC-OCP} is:
\begin{equation}\label{eq: Decentralized_SC-OCP}
    \begin{aligned}
    &V_d(t, o(t)) = \min_{\ctrlseq_d}\; \int_t^{T} l_d(o(s))\, ds + \phi_d(o(T)) \\
    \text{s.t.} ~ & \dot{o}(s) = f_d(o(s), u_d(s)),
    ~ g_d(o(s)) \leq 0 ~~ \forall s \in [t, T]
    \end{aligned}
\end{equation}
Here, $l_d$ and $\phi_d$ denote the local running and terminal costs, $f_d$ denotes the local dynamics, and $g_d$ enforces collision avoidance among the agents contained in the observation. By fixing the observation size to $n+1$ agents, the decentralized formulation becomes independent of the dimension of the global state space, enabling a scalable approximation of the original SC-OCP.

\subsection{Epigraph Reformulation to Solve Decentralized SC-OCP}\label{subsec: epigraph}

Directly solving the decentralized SC-OCP in~\eqref{eq: Decentralized_SC-OCP} still remains challenging due to the presence of hard state constraints. To encode these objectives within a single value function, we introduce an auxiliary variable $z\in\mathbb{R}^+$ that represents an admissible cost budget. Inspired by epigraph formulations for constrained optimal control~\cite{altarovici2013general}, this allows the original constrained optimization problem to be recast as the search for the smallest admissible value of $z$.
Specifically, we define a new auxiliary value function $\hat{V}$:
\begin{equation}\label{eq: aux_value_func}
    V_d(t, o(t)) = \min_{z \in \mathbb{R}^+}  z \quad \text{s.t.} \quad \hat{V}_d(t, o, z) \leq 0,
\end{equation}
Following the framework of~\cite{altarovici2013general}, this auxiliary value function is given by:
\begin{equation}\label{eq: aux_vfunc_def}
    \hat{V}_d(t, o(t), z) = \min_{{\ctrlseq}_d} \max \left\{ C(t, o(t), \ctrlseq_d) - z,\; \max_{s \in [t, T]} g_d(o(s)) \right\}.
\end{equation}

The auxiliary value function evaluates whether a given cost budget $z$ can be achieved while satisfying the safety constraints over the entire horizon. Consequently, solving~\eqref{eq: aux_value_func} amounts to identifying the minimum feasible cost that remains compatible with safe operation.

To further enable the application of dynamic programming principles, we treat the auxiliary variable $z$ as a state variable with the simple dynamics $\dot{z}(t) = -l(o(t))$. This signifies that the admissible cost bound $z$ is depleted by the stage cost $l(o)$ along the system's trajectory. This yields the following augmented system dynamics:
\begin{equation}
    \dot{\hat{o}} = \hat{f}_d(t, \hat{o}, u) := 
    \begin{bmatrix}
        f_d(t, o, u) \\
        -l(o)
    \end{bmatrix},
\end{equation}
where $\hat{o} := [o, z]^\top \in \mathcal{X} \times \mathbb{R}$ is the augmented state.

Under standard assumptions from~\cite{altarovici2013general}, the resulting safety-performance value function is characterized as the unique continuous viscosity solution to the Hamilton-Jacobi-Bellman (HJB) partial differential equation:
\begin{equation}\label{eq: coopt_pde}
    \min\left(-\partial_t \hat{V}_d - \min_{\ctrlseq_d} \left\langle \nabla_{\hat{o}} \hat{V}_d,\, \hat{f}_d(\hat{o}, u) \right\rangle,\; \hat{V}_d - g_d(o)\right) = 0,
\end{equation}
for all $t \in [0,T)$ and $\hat{o} \in \mathcal{X} \times \mathbb{R}$, with the terminal condition:
\begin{equation}\label{eq: terminal_condition}
    \hat{V}_d(T,\hat{o}) = \max\left( \phi_d(o) - z,\; g_d(o) \right).
\end{equation}

For notational simplicity, we henceforth drop the subscript $d$ and denote the decentralized auxiliary value function by $\hat V(t,\hat o)$.

%% file: 3_method.tex
%
The solution to the decentralized SC-OCP in Equation~\eqref{eq: Decentralized_SC-OCP} hinges on the computation of the optimal value function $V_d$ that minimizes cost under safety constraints. Our approach to obtaining this function proceeds in two primary stages as illustrated in Figure~\ref{fig:algorithm}: an offline learning phase and an online deployment phase. First, we learn the auxiliary value function $\hat{V}$ using a physics-informed machine learning framework. Then, $V_d$ is obtained from $\hat{V}$ using \eqref{eq: aux_value_func}. For online decentralized deployment, a safety-aware clustering strategy is employed to determine the appropriate neighbours for each agent. The control policy for each agent is then derived based on $V_d$. We now discuss each step in detail.

\subsection{Training the Auxiliary Value Function ($\hat{V}$)}
The auxiliary value function $\hat{V}$ is characterized by the HJB-PDE in \eqref{eq: coopt_pde} (Section \ref{subsec: epigraph}). Traditional numerical methods for solving such PDEs rely on discretizing the state space over a grid \cite{mitchell2004toolbox, pythonhjtoolbox}. While accurate for low-dimensional systems, these methods are susceptible to the curse of dimensionality, as their computational cost scales exponentially with the number of states. To overcome this limitation, we leverage a physics-informed neural network (PINN) framework that uses PDE residuals to learn the value function and has shown promising results in solving high-dimensional HJB PDEs \cite{9561949}.

\vspace{0.2em}
\noindent \textit{Auxiliary Value Function Parameterization:} We approximate the auxiliary value function $\hat{V}(t, \hat{o})$ using a neural network, with parameters $\theta$. A critical requirement is that the solution must satisfy the terminal boundary condition to adhere to the problem's safety constraints. To enforce this \textit{exactly}, we structure our network output as:
\begin{equation*}
    \hat{V}_\theta(t, \hat{o}) = \max\left( \phi_d(o) - z,\; g_d(o) \right) + (T - t) \cdot R_\theta(t, \hat{o}),
\end{equation*}
where the first term encodes the terminal condition in \eqref{eq: terminal_condition}, and the neural network, denoted $R_\theta(t, \hat{o})$, learns the residual evolution of the value function over time. This formulation, inspired by \cite{11127972}, guarantees that $\hat{V}_\theta(T, \hat{o}) = \max\left( \phi_d(o) - z,\; g_d(o) \right)$ for any state $\hat{o}$, irrespective of the network's output $R_\theta$. 
There are two key advantages to the proposed structure of $\hat{V}_\theta$: (a) it eliminates the need to explicitly learn a complex boundary condition by hard-coding it into the network's forward pass, and (b) it reduces the learning problem to minimizing a single HJB-derived loss function (as we discuss next), thereby removing the necessity for a manually-tuned loss weighting scheme.

\vspace{0.2em}
\noindent \textit{Loss Function and Training Scheme:} The parameters $\theta$ of the network $R_\theta$ are learned by minimizing a loss function that penalizes the HJB PDE residual errors. Specifically, the loss function to learn the NN parameters is:
\begin{equation}\label{eq: loss}
\begin{aligned}
&\mathcal{L}\left(t_k, \hat{o}_k | \theta\right)=\mathcal{L}_{pde}\left(t_k, \hat{o}_k | \theta\right) \\
&=\| \min \left\{-\partial_t \learnedauxvfunc\left(t_k, \hat{o}_k\right)- H(t_k,\hat{o}_k),\right. \left.\learnedauxvfunc\left(t_k, \hat{o}_k\right) - g_d\left(o_k\right)\right\} \| ,\\
&=\| \min \left\{(t_k-T) \partial_t R_\theta\left(\hat{o}_k, t_k\right) + R_\theta\left(\hat{o}_k, t_k\right)-H\left(\hat{o}_k, t_k\right),\right. \\
& \qquad \qquad \left.V_\theta\left(\hat{o}_k, t_k\right) - g_d\left(o_k\right)\right\} \|,
\end{aligned}
\end{equation}
where, $H(t,\hat{o}_i) = \min_{u\in \mathcal{U}} \langle \nabla V_{\theta}(\hat{o}_i,t),\hat{f}_d(\hat{o}_i,u) \rangle$. Typically, PINNs incorporate an additional loss term to enforce boundary conditions. In contrast, due to the structure of our formulation of $\hat{V}_{\theta}$, the boundary conditions are satisfied exactly. Consequently, the optimization reduces to minimizing a single loss term $\mathcal{L}_{pde}$, eliminating the need for auxiliary loss terms and the hyperparameters required for their weighting.

\vspace{0.2em}
\noindent \textit{Curriculum Training:} A key challenge in solving the HJB-PDE is its backward-in-time evolution; the solution at time $t$ depends on the future time $t + \Delta t$. To manage this complexity during training, we employ a curriculum learning strategy similar to DeepReach~\cite{9561949}. Training begins by sampling time points near the terminal time $T$ and progressively expanding the interval backward until it covers the entire horizon $[0, T]$, whereas the states are sampled uniformly across the state space at each training iteration.
This allows the network to first learn the value function accurately at terminal time, then propagate the solution backward in time under the PDE dynamics, yielding the auxiliary value function $\hat{V}_\theta$. We refer interested readers to \cite{9561949, 11127972} for more details on the curriculum training scheme. 

\begin{figure*}[t]
    \centering
    \includegraphics[width=1.0\textwidth]{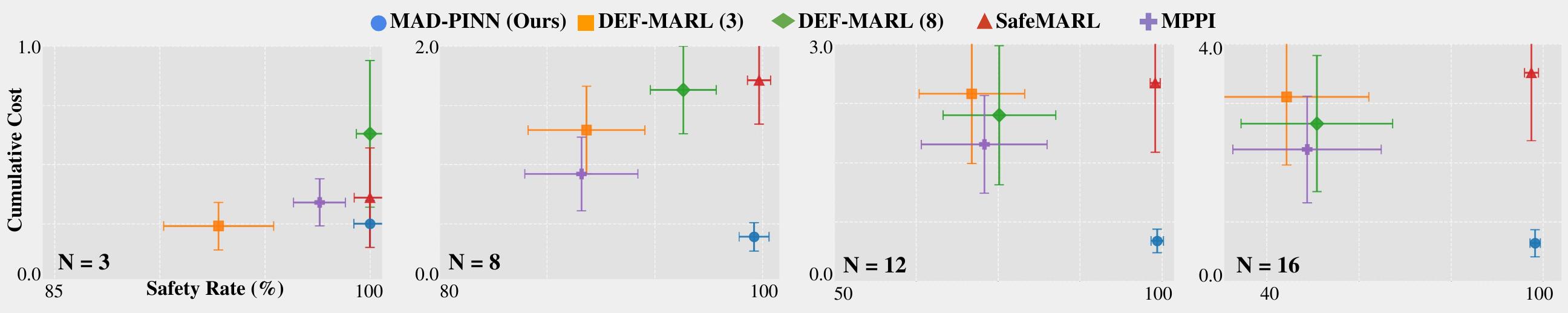}
    \caption{Comparison of cumulative costs and safety rates across testing environments with $3$, $8$, $12$, and $16$ agents. Our method consistently appears in the bottom-right region, indicating \textbf{superior safety-performance co-optimization relative to all baselines}. Moreover, its degradation in performance and safety with increasing agent count is minimal, \textbf{demonstrating better scalability compared to the baselines}.}
    \label{fig:Cost_safety_plot}
    \vspace{-2em}
\end{figure*}

\subsection{Neighbour Selection Strategy}
During online deployment, each agent must efficiently identify the $n$ neighbors most relevant to its policy execution. Rather than identifying neighbors based on physical distance or treating all nearby agents equally, we select the $n$ neighbors most critical to maintaining safety.
%
To determine which neighbors pose the greatest risk, we employ a principled criterion derived from Hamilton–Jacobi (HJ) reachability analysis. HJ reachability provides a rigorous way to quantify the likelihood of safety conflicts, enabling each agent to prioritize interactions that most directly impact safe operation. Specifically, while $\hat{V}$ and $V_d$ encode cost and safety constraints in the SC-OCP, we additionally compute a pairwise safety value function $V_s$ using HJ reachability to guide neighbor selection.

HJ Reachability\cite{mitchell2005time, lygeros2004reachability} characterizes the set of states from which the system can be driven into a failure set. Let $\psi:\mathbb{R}^n \to \mathbb{R}$ be a Lipschitz-continuous function whose sub-zero level set $\Psi=\{x:\psi(x)\leq0\}$ represents collision states between the two agents. The corresponding safety value function is given by:
\begin{equation}
  V_s(x,t) = \sup_{u(\cdot)} \; \min_{\tau \in [t,T]} \psi(\xi^u_{x,t}(\tau)),
\end{equation}
where $\xi^u_{x,t}(\cdot)$ is the system trajectory. 
Intuitively, $V_s(x,t)$ measures how close the two agents are to a collision, even under optimal control. The unsafe set -- the set of states from which collision is inevitable -- is precisely the sub-zero level set of $V_s$.
The value function satisfies the Hamilton-Jacobi-Bellman Variational Inequality (HJB-VI):
\begin{equation}
\begin{gathered}
\label{eq: HJI-VI}
     \min \{\partial_t V_s(x,t) + H_s(x,t), \psi(x) - V_s(x,t) \} = 0, \\
     V_s(x,T) = \psi(x),\\
    H_s(x,t) = \max_{u \in \mathcal{U}} \langle \nabla V_s(x,t) \; , \; f_s(x,u) \rangle, 
\end{gathered}
\end{equation}
where $H_s$ is the Hamiltonian corresponding to the safety value function and $f_s$ encodes the pairwise dynamics between agents. 

\vspace{0.2em}
\noindent \textit{{Learning the Pairwise Safety Value Function ($V_s$):}} 
We approximate $V_s$ using DeepReach~\cite{9561949, 11127972}.
To characterize pairwise safety interactions, we consider each agent $i$ in relation to another agent $j$. The target function is defined as $\psi = d_{ij} - r$, where $d_{ij}$ denotes the distance between the two agents and $r$ is the prescribed collision radius. By design, $\psi \leq 0$ corresponds to states in which agents $i$ and $j$ are in collision, thereby providing a natural safety signal. Training in this manner yields a value function that quantifies the relative safety risk posed by one agent to another.
Note that since the agents are homogenous, we only need to compute $V_s$ once for a generic pair of agents. 

\vspace{0.2em}
\noindent \textit{Neighbour Selection:} 
For agent $i$, the value $(V_s)_{ij}$ represents the degree of risk posed by agent $j$. Smaller values correspond to higher collision likelihoods. Specifically, if $(V_s)_{ij} < (V_s)_{ik}$, then agent $j$ poses a higher safety risk to agent $i$ compared to agent $k$, and thus should be prioritized in $i$'s decision-making process. Hence, to select its $n$ neighbours, agent $i$ computes $(V_s)_{ij}$ for all agents $j$, ranks them, and selects the $n$ agents with the lowest values.
More importantly, $V_s$ quantifies safety beyond physical proximity alone by accounting for the full state of the agents, such as their relative velocity and heading.
This process ensures that each agent focuses its decision-making on the most safety-critical interactions, while still keeping the neighbourhood size fixed for computational tractability.

\subsection{Policy Synthesis}\label{subsec: policy_synthesis}
Once the neighbour set is determined, each agent synthesizes its policy by solving the optimization problem in \eqref{eq: aux_value_func}.
To enforce safety, we set $V_d(t,x)=+\infty$ whenever $\learnedauxvfunc(t,x,z)>0$, since such states are unsafe and violate the safety constraint. For the remaining states, the optimization is solved via binary search over $z$. The resulting state-feedback policy for agent $i$, $\learnedpolicy^i:\mathcal{X}\times[t,T)\to\mathcal{U}$, is given by
\begin{equation*} \label{eq: optimal_pi}
\learnedpolicy^i(t,x)=\arg\min_{\ctrlseq_d}\langle\nabla\learnedauxvfunc(t,\hat{o}^*),\;\hat{f}_d(\hat{o}^*,\ctrlseq_d)\rangle,
\end{equation*}
where $\hat{o}^*=[o,z^*]^T$ is the augmented state corresponding to the optimal $z^*$.  

Since interaction structures in multi-agent navigation evolve dynamically, the neighbour set $\mathcal{N}_i$ is updated online, motivating a \textit{receding-horizon execution} of the policy. 
This ensures that policy continuously adapts by incorporating updated interaction information and anticipating future conflicts. 
In addition, frequent re-planning further enhances robustness to model mismatch, sensor noise, and external disturbances, enabling reliable decentralized navigation. 
Finally, this framework naturally extends to long-horizon tasks by repeatedly solving the SC-OCP over shorter horizons, allowing the proposed framework to co-optimize safety and performance in a practical and computationally efficient manner for real-world autonomous systems.

\subsection{Computational Complexity Analysis}
The online execution of MAD-PINN consists of two stages: (i) safety-aware neighbor selection and (ii) policy inference. Since both the pairwise safety value function $V_s$ and the auxiliary value function $\hat{V}$ are trained offline, their online evaluation incurs a fixed inference cost independent of the number of agents.
Consider a single agent $i$. During neighbor selection, agent $i$ evaluates the pairwise safety value function $V_s$ with respect to other agents in the environment and ranks the resulting safety values to identify the $n$ most safety-critical neighbors. Let $T_s$ denote the inference time of the safety value network. Since $N-1$ pairwise interactions must be evaluated, the computational cost of neighbor selection for agent $i$ is $(N-1)T_s$. Once the neighbor set has been determined, agent $i$
computes its control action using the learned
auxiliary value function $\hat{V}$. Since the neighborhood size $n$ is fixed, the input dimension of $\hat{V}$ is independent of the total number of agents $N$. Therefore, the policy inference stage requires only a fixed number of evaluations of $\hat{V}$, resulting in an $\mathcal{O}(1)$ computational cost with respect to $N$. Let $T_v$ denote the corresponding inference cost. Combining both stages, the online computational cost for a single agent is: $T_i = (N-1)T_s + T_v$. Since $T_s$ and $T_v$ are fixed constants determined by the network architectures, the per-agent online complexity scales linearly with the number of agents, yielding an overall complexity of $\mathcal{O}(N)$. Importantly, the pairwise safety evaluations performed during neighbor selection are independent and correspond to repeated evaluations of the same network $V_s$. As a result, they can be executed as a single batched inference operation. While batching does not change the asymptotic $\mathcal{O}(N)$ complexity, it reduces the practical runtime through parallel evaluation.

%% file: 4_experiments.tex
\begin{figure*}[t]
    \centering
    \includegraphics[width=1.0\textwidth]{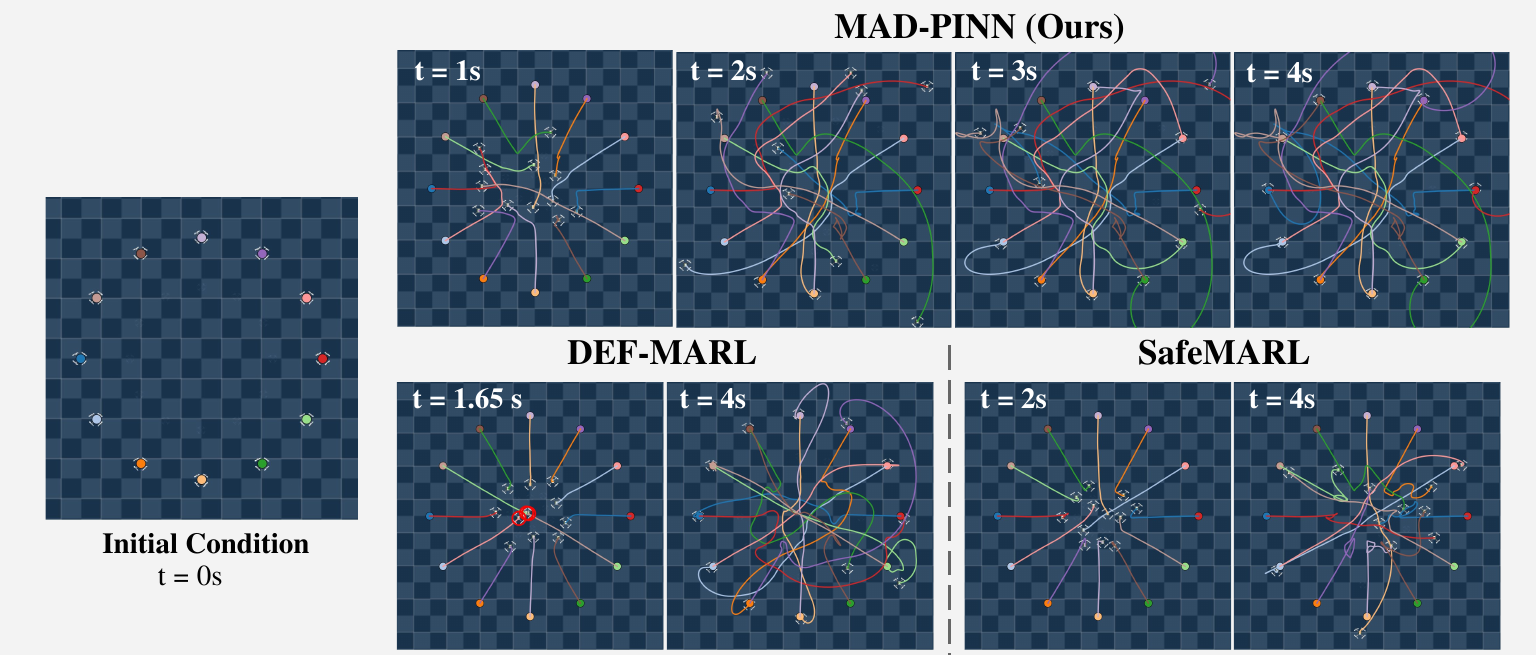}
    \caption{Multi-agent navigation with Crazyflie quadrotors in the MuJoCo physics simulator.
    Snapshots of multi-agent navigation trajectories at different times using MAD-PINN and baselines. Agents are represented as drones with radius $R$, indicating the minimum safe distance they must maintain from each other. The dots mark their respective goals. MAD-PINN trajectories show that agents proactively \textbf{maintain long-horizon safety} by adjusting their paths to avoid close encounters, rather than enforcing safety reactively (SafeMARL), which could lead to suboptimal behaviors. Furthermore, while less conservative than SafeMARL, \textbf{DEF-MARL experiences collisions} (red circles), limiting its ability to co-optimize safety and performance.}
    \label{fig:Trajectory}
\end{figure*}

\begin{table*}[h!]
\centering
\resizebox{1.0\textwidth}{!}{\begin{tabular}{lcccccccc}
\toprule
& \multicolumn{2}{c}{3 Agents} & \multicolumn{2}{c}{8 Agents} & \multicolumn{2}{c}{12 Agents} & \multicolumn{2}{c}{16 Agents} \\
\cmidrule(lr){2-3}\cmidrule(lr){4-5}\cmidrule(lr){6-7}\cmidrule(lr){8-9}
Method & Safety & Safe Sc. & Safety & Safe Sc. & Safety & Safe Sc. & Safety & Safe Sc. \\
\midrule
Ours              & 100\% $\pm$ 0.0\% & 100\% $\pm$ 0.0\% & 99.5\% $\pm$ 0.4\% & 98\% $\pm$ 1.2\%  & 99.3\% $\pm$ 0.6\% & 96\% $\pm$ 1.8\%  & 98.3\% $\pm$ 0.8\% & 86\% $\pm$ 2.7\% \\
DEF-MARL (3)      & 94\% $\pm$ 2.8\%  & 91\% $\pm$ 3.4\%  & 89.8\% $\pm$ 2.5\% & 63\% $\pm$ 4.2\%  & 71\% $\pm$ 3.8\%   & 31\% $\pm$ 4.8\%  & 44.3\% $\pm$ 3.4\% & 7\% $\pm$ 1.5\%  \\
DEF-MARL (8)      & 100\% $\pm$ 0.0\%    & 100\% $\pm$ 0.0\%    & 95.4\% $\pm$ 2.1\% & 82\% $\pm$ 3.2\%  & 75.2\% $\pm$ 3.5\% & 37\% $\pm$ 4.5\%  & 50.9\% $\pm$ 2.9\% & 13\% $\pm$ 2.1\% \\
SafeMARL  & 100\% $\pm$ 0.0\% & 100\% $\pm$ 0.0\% & 99.8\% $\pm$ 0.2\% & 99\% $\pm$ 0.8\%  & 99\% $\pm$ 0.5\%   & 96\% $\pm$ 1.7\%  & 97.5\% $\pm$ 1.0\% & 90\% $\pm$ 2.2\% \\
MPPI              & 98\% $\pm$ 1.1\%  & 97\% $\pm$ 1.4\%  & 89.5\% $\pm$ 2.8\% & 72\% $\pm$ 3.5\%  & 72.9\% $\pm$ 3.7\% & 35\% $\pm$ 4.9\%  & 48.8\% $\pm$ 3.6\% & 9\% $\pm$ 2.4\%  \\
\bottomrule
\end{tabular}}
\caption{Average safety rates and safe-scenario percentages across varying numbers of agents. Our method maintains consistently high values, indicating collision-free execution of every agent for nearly all initial configurations and demonstrating \textbf{effective safety-performance co-optimization at the agent and system level}.}
\label{tab:safety_scaling}
\vspace{-1em}
\end{table*}

\begin{figure}[ht]
    \centering
    \includegraphics[width=0.65\columnwidth]{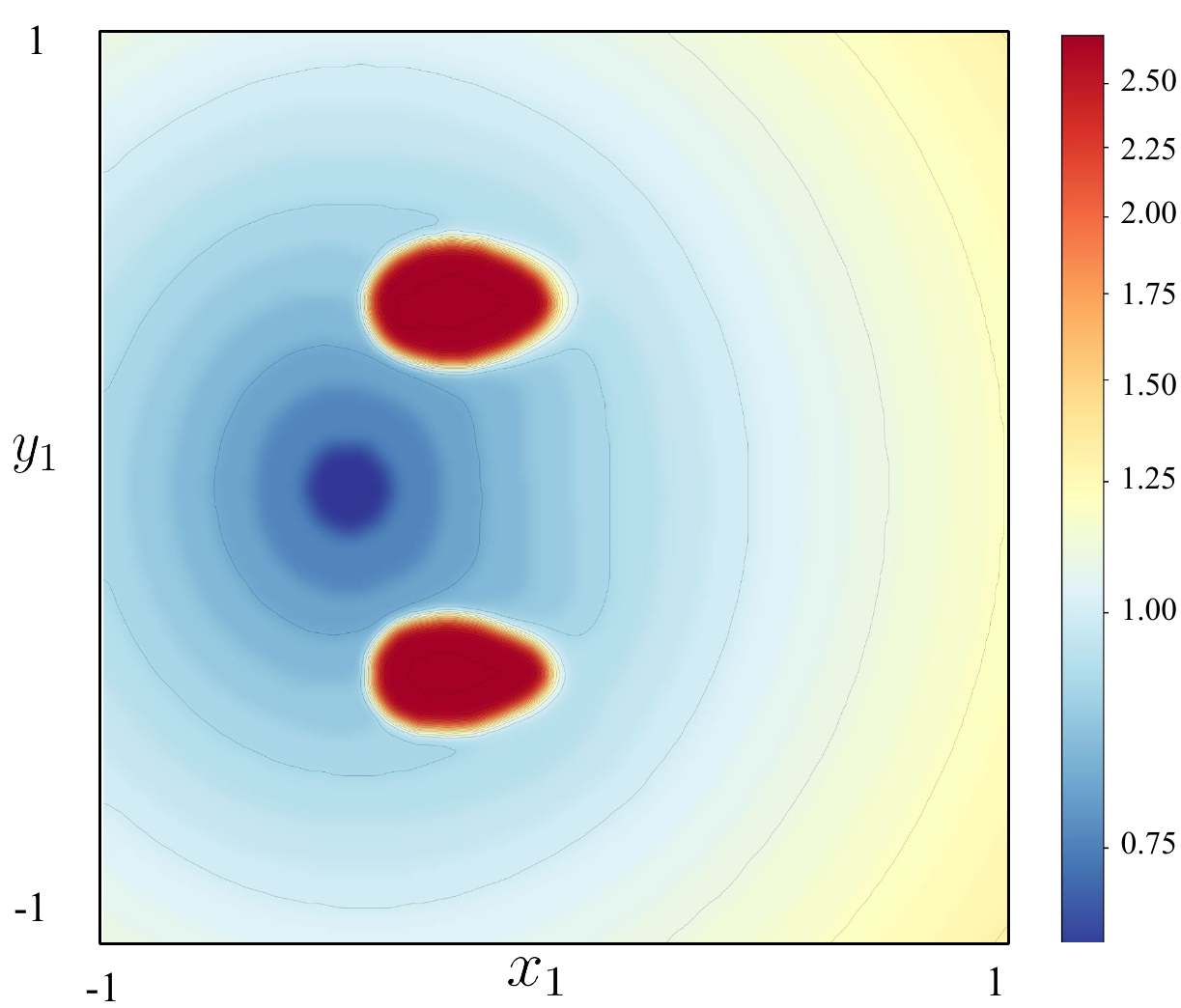}
    \caption{\textbf{Heatmap of the learned value function with respect to the ego agent’s position coordinates.} The other two agents are at $[-0.3,\,0.4]$ and $[-0.3,\,-0.4]$, both moving with velocity $[-1,\,0]$, while the ego agent moves with velocity $[1,\,0]$ toward its goal at $[-0.5,\,0]$.}
    \label{fig:Val_Func}
\end{figure}

The goal of our experiments is to assess the effectiveness of the proposed framework in (i) co-optimizing safety and performance in multi-agent systems, (ii) scaling to larger environments with higher numbers of agents, and (iii) validating the proposed safety-aware neighbour selection strategy. 

\subsection{Baselines}
To thoroughly assess our method, we benchmark it against baselines that represent the spectrum of safety integration techniques: $1)$ \emph{Constrained Policy Synthesis:} \textbf{DEF-MARL}~\cite{zhang2025defmarl}, which leverages multi-agent RL (MARL) to solve the discrete-time epigraph reformulation of the SC-OCP to synthesize safe and optimal policies; $2)$ \emph{Safety Filtering:}
\textbf{SafeMARL}~\cite{choi_aloor_li_2025safemarl}, which uses a control barrier value function (CBVF)-based safety filter to provide safety for a nominal MARL policy; and $3)$ \emph{Soft-Constrained Optimization:} \textbf{MPPI}~\cite{8558663}, a sampling-based MPC method to solve \eqref{eq: Decentralized_SC-OCP} that penalizes constraint violations in its cost function using a Lagrangian approach.

\subsection{Evaluation Metrics}
To evaluate the trade-off between performance and safety, we use the following metrics: \textbf{(1) Cumulative Cost:} The total cost $\int_0^T l(x(s))ds + \varphi(x(T))$ accumulated along safe trajectories.
\textbf{(2) Safety Rate:} The fraction of agents that remain collision-free for the entire horizon, quantifying per-agent safety. \textbf{(3) Safe Scenarios:} The fraction of scenarios in which all agents remain collision-free for the entire horizon, reflecting collective safety across the multi-agent system.  

\subsection{Experimental Setup}
We study a planar multi-agent drone navigation problem involving \textbf{Crazyflie Quadrotors in the MuJoCo Physics Simulator}. We approximate each agent using double-integrator dynamics with state $\mathbf{s}_i = [x, y, v_x, v_y]^\top$, constrained to $(x, y) \in [-1,1]^2$ and $(v_x, v_y) \in [-4,4]^2$. The control input is $\mathbf{u}_i = [a_x, a_y]^\top \in [-4,4]^2$, representing acceleration. Each agent is assigned a parameterized goal $(x_g, y_g)$, and its dynamics are given by $\dot{x}=v_x$, $\dot{y}=v_y$, $\dot{v}_x=a_x$, $\dot{v}_y=a_y$. 
The learned acceleration command $a_i$ is integrated over one policy step to a position–velocity setpoint ($p_d,v_d$), which is forwarded to the onboard cascaded position–attitude controller (identical to the Crazyflie firmware loop \cite{5980409}) that synthesizes the per-rotor RPMs; we verify the resulting closed-loop behavior on the Crazyflie 2.X model in MuJoCo-Drones-Gym\cite{tayal2026mujoco}.
Agents must reach their goals while maintaining a safety distance $r=0.1$. The running cost for training the decentralized value function is:
\begin{equation}
    l_d = \sum_{j=1}^{|\mathcal{N}_i|} \|(x^j(t), y^j(t))^\top - (x^j_g(t), y^j_g(t))^\top\|,
\end{equation}
where $\mathcal{N}_i$ denotes the neighbor set. We train the auxiliary value function with 3 agents (so $|\mathcal{N}_i|=2$), a time horizon of $0.2$s, 
and deploy the same value function for all agents across all environments. 
The residual component of the auxiliary value function $R_\theta$ is approximated as a multi-layer perceptron (MLP) with three hidden layers of 256 neurons each, using sine activation functions. The network is trained with the Adam optimizer at a learning rate of $2 \times 10^{-5}$.
We evaluate all algorithms over $100$ distinct initial conditions across $5$ seeds.

Figure~\ref{fig:Val_Func} illustrates the heatmap of the learned value function as a function of the ego agent’s position coordinates. Regions of high value (red) correspond to unsafe states overlapping with the positions of other agents, thus encoding safety constraints. Conversely, regions of low value (blue) are concentrated around the goal, reflecting the task objective. The smooth transition between these regions demonstrates that the learned value function simultaneously captures both safety constraints and task objectives, enabling safety-performance co-optimization.

\subsection{Results}
%
\begin{table}[t]
\centering
\begin{tabular}{lccc}
\toprule
Number of Agents & Cumulative Cost & Safety Rate & Safe Scenarios \\
\midrule
64 & 1.09 & 100\% & 100\%\\
128 & 1.68 & 98.75\% & 95\% \\
256 & 2.73 & 96.25\% & 85\% \\ 
\bottomrule
\end{tabular}
\caption{Effect of increasing agent count and environment size. Our method \textbf{co-optimizes safety and performance even in large multi-agent environments}, despite training on smaller environments with fewer agents.}
\label{tab:ours_scaling}
\vspace{-2.1em}
\end{table}
\noindent \textbf{1) Effectiveness in Co-optimizing Safety and Performance:} To evaluate safety-performance co-optimization, we compare our method against all baselines on scenarios with $3$, $8$, $12$, and $16$ agents while keeping the environment size fixed, thereby increasing agent density. As shown in Figure~\ref{fig:Cost_safety_plot}, the proposed method consistently achieves the best safety-performance trade-off, attaining the lowest cumulative cost while maintaining near-perfect safety rates across all settings. In contrast, DEF-MARL performs competitively on the agent counts used during training but exhibits a noticeable degradation in both safety and performance as density increases, highlighting the limited generalization of RL-based co-optimization approaches. SafeMARL maintains high safety but incurs substantially higher costs due to conservative behavior, while MPPI performs poorly on both metrics because safety is enforced only through soft penalties. Table~\ref{tab:safety_scaling} further evaluates scalability using the more stringent \textit{safe scenarios} metric, which requires collision-free execution by all agents over the entire horizon. The proposed method consistently achieves a high percentage of safe scenarios across all agent counts, whereas DEF-MARL and MPPI deteriorate rapidly with increasing density. These results demonstrate that effective safety-performance co-optimization can be achieved using only local observations, with minimal performance degradation when scaling from $3$ to $16$ agents. Figure~\ref{fig:Trajectory} further corroborates these findings, showing that all agents safely reach their goals without collisions, whereas the baselines either exhibit overly conservative behavior or fail to maintain long-horizon safety.

\noindent \textbf{2) Scalability with Agent Count and Environment Size:} To evaluate the scalability of our approach, we conduct experiments with $20$ distinct initial conditions and a substantially larger number of agents, namely $64$, $128$, and $256$. In addition, the environment size is increased from $[-1,1]^2$ to $[-4,4]^2$ to test the method in a more extensive and populated setting compared to the training environment. As shown in Table~\ref{tab:ours_scaling}, the proposed method maintains high safety rates and safe-scenario percentages even with $256$ agents, demonstrating that decentralized safety-performance co-optimization remains effective at scale. While cumulative costs increase due to the larger environment and longer travel distances, the increase is approximately proportional to the environment size, suggesting that our method preserves the same level of performance as in Table~\ref{tab:safety_scaling}. These results confirm the ability of our framework to address large-scale multi-agent problems by training on local observations and deploying the same policy independently across hundreds of agents. This also highlights the practicality of our framework for deployment in large-scale multi-agent systems.

\begin{table}[t]
\centering
\begin{tabular}{lccc}
\toprule
Method & Cumulative Cost & Safety Rate & Safe Scenarios \\
\midrule
Value-based (Ours) & 0.51 & 99.33\% & 96\% \\
Nearest & 0.82 & 83\% & 45\% \\
Random & 1.55 & 33\% & 4\% \\
\bottomrule
\end{tabular}
\caption{Ablation study on the impact of the proposed neighbor selection strategy. By effectively \textbf{capturing safety-critical inter-agent interactions among neighbors}, it enables improved safety–performance co-optimization.}
\label{tab:ablation}
\vspace{-2.0em}
\end{table}

\noindent \textbf{3) Effectiveness of Neighbor Selection:} We conduct an ablation study to evaluate the effect of the proposed HJ Reachability-based neighbor selection strategy against two alternatives: (i) selecting the two nearest neighbors and (ii) randomly selecting two neighbors within an observation radius. All methods are tested on the same 12-agent environment used in Table~\ref{tab:safety_scaling}. As shown in Table~\ref{tab:ablation}, the proposed neighbor selection strategy achieves the highest safety rates and lowest costs, demonstrating that the pairwise safety value function $V_s$ effectively identifies the most safety-critical interactions. In contrast, distance-based selection ignores important dynamic information, such as relative velocities, leading to reduced safety, while random selection performs worst due to the lack of any interaction-aware prioritization. These results highlight the necessity of a principled neighbor selection framework that accounts for critical inter-agent interaction information to ensure reliable decentralized multi-agent navigation.

\begin{table}[h]
\centering
\begin{tabular}{lccc}
\toprule
Update Steps ($k$) & Cumulative Cost & Safety Rate & Safe Scenarios \\
\midrule
1 & 0.49 & 99.99\% & 99\% \\
5~(Our setting) & 0.51 & 99.33\% & 96\% \\
10 & 0.52 & 98.00\% & 89\% \\
20 & 0.56 & 92.67\% & 67\% \\
50 & 0.59 & 66.83\% & 11\% \\
100 & 0.65 & 36.75\% & 7\% \\
500 & 0.69 & 21.17\% & 4\% \\
$\infty$~(No update) & 0.78 & 15.92\% & 2\% \\
\bottomrule
\end{tabular}
\caption{Ablation study on the impact of the proposed neighbor update frequencies. \textbf{Safety degrades significantly as updates become infrequent}, highlighting the importance of periodically recomputing safety-critical neighbors.}
\label{tab:update_ablation}
\vspace{-1em}
\end{table}
\noindent \textbf{4) Effect of Neighbor Update Frequencies:} Finally, we also conduct an ablation study to evaluate the effect of neighbor update frequencies on the safety and performance metrics. All methods are tested on the same 12-agent environment used in Table~\ref{tab:safety_scaling}. The simulation timestep is $0.0025~\mathrm{s}$, and neighbor updates are performed every $k$ simulation steps. The neighbor selection process takes $1.2~\mathrm{ms}$ per agent. 
Table~\ref{tab:update_ablation} shows that more frequent updates improve both safety and performance by enabling agents to react to changing local interactions. While updating every timestep yields the best results, the proposed setting of $5$ steps ($0.0125~\mathrm{s}$) achieves comparable safety and performance while reducing the computational overhead associated with repeated neighbor selection. As the update interval increases, safety degrades much faster than performance, indicating that stale neighborhood information primarily affects collision avoidance. In the extreme case of no updates ($k=\infty$), safety deteriorates significantly, highlighting the importance of dynamic neighbor selection. Overall, the results demonstrate that dynamic neighbor updates are critical for maintaining safety, with safety metrics exhibiting substantially greater sensitivity to update frequency than task performance.


%% file: 5_conclusion.tex
We presented a physics-informed learning framework for scalable multi-agent safe and optimal control. By reformulating the large-scale SC-OCP into a decentralized problem with fixed observation size and introducing a reachability-based neighbour selection strategy, the proposed approach mitigates the curse of dimensionality while accounting for the most safety-critical interactions. Experimental results demonstrated strong safety-performance co-optimization and favorable scalability across a wide range of agent densities. Future work will focus on extending the framework to heterogeneous systems, incorporating model uncertainty, and evaluating performance on real-world robotic platforms. We also plan to quantify the approximation error of the learned auxiliary value function, e.g., via conformal prediction~\cite{pmlr-v242-lin24a, tayal2025cp}, and investigate its effect on safety and performance.